\documentclass[11pt,a4paper]{article}
\usepackage[hyperref]{acl2019}

\usepackage[utf8]{inputenc}
\usepackage{xspace}
\usepackage{times}
\usepackage{latexsym}
\usepackage{microtype}
\usepackage{float}

\usepackage{booktabs}
\usepackage{multirow}
\usepackage{multicol}

\usepackage{tikz}
\usepackage{pgfplots}
\pgfplotsset{compat=1.11}

\usepackage{todonotes}
\usepackage{url}
\usepackage{lipsum}

\usepackage{bibentry}
\usepackage{subfiles}

\usepackage{import}

\usepackage{amsmath}

\newcommand\bleu{\textsc{BLEU}\xspace}
\newcommand\sacrebleu{\textsc{SacreBLEU}\xspace}

\usepackage{lettrine}

\aclfinalcopy 

\title{The University of Edinburgh’s Submissions\linebreak to the WMT19 News Translation Task}

\author{Rachel Bawden \quad Nikolay Bogoychev \quad Ulrich Germann \quad Roman Grundkiewicz \\
\textbf{Faheem Kirefu} \quad \textbf{Antonio Valerio Miceli Barone} \quad \textbf{Alexandra Birch}  \\
\\
  School of Informatics, University of Edinburgh, Scotland \\
  \texttt{rachel.bawden@ed.ac.uk}} 

\date{}

\begin{document}
\maketitle
\begin{abstract}
  The University of Edinburgh participated in the WMT19 Shared Task on News Translation in six language directions: 
  English$\leftrightarrow$Gujarati, English$\leftrightarrow$Chinese, German$\rightarrow$English, and\linebreak English$\rightarrow$Czech.
  For all translation directions, we created or used back-translations of monolingual data in the target language as additional synthetic training data.
  For English$\leftrightarrow$Gujarati, we also explored semi-supervised MT with cross-lingual language model pre-training, and translation pivoting through Hindi.
  For translation to and from Chinese, we investigated character-based tokenisation vs. sub-word segmentation of Chinese text. For German$\rightarrow$English, we studied the impact of vast amounts of back-translated training data on translation quality, gaining a few additional insights over \citet{edunov_understanding_2018}.
  For English$\rightarrow$Czech, we compared different pre-processing and tokenisation regimes.
\end{abstract}

\section{Introduction}
The University of Edinburgh participated in the WMT19 Shared Task on News Translation in six language directions:
English-Gujarati (EN$\leftrightarrow$GU), English-Chinese (EN$\leftrightarrow$ZH), German-English (DE$\rightarrow$EN) and English-Czech (EN$\rightarrow$CS). 
All our systems are neural machine translation (NMT) systems 
trained in constrained data conditions with the Marian\footnote{\url{https://marian-nmt.github.io}} toolkit \citep{junczys-dowmunt_marian:_2018}.
The different language pairs pose very different challenges, 
due to the characteristics of the languages involved and arguably more importantly, due to the amount of  training data available.

\paragraph{Pre-processing}
For EN$\leftrightarrow$ZH, we investigate character-level pre-processing for Chinese compared with subword segmentation. For EN$\rightarrow$CS, we show that it is possible in high resource settings to simplify pre-processing by removing steps.

\paragraph{Exploiting non-parallel resources}
For all language directions, we create additional, synthetic parallel training data. 
For the high resource language pairs, we look at ways of effectively using large quantities of backtranslated data. 
For example, for DE$\rightarrow$EN, we investigated the most effective way of combining genuine parallel data with larger quantities of synthetic parallel data and for CS$\rightarrow$EN, we filter backtranslated data by re-scoring translations using the MT model for the opposite direction. The challenge for our low resource pair, EN$\leftrightarrow$GU, is producing sufficiently good models for back-translation, which we achieve by training semi-supervised MT models with cross-lingual language model pre-training~\citep{lample_cross-lingual_2019}. We use the same technique to translate additional data from a related language, Hindi.

\paragraph{NMT Training settings}
In all experiments, we test state-of-the-art training techniques, including using ultra-large mini-batches for DE$\rightarrow$EN and EN$\leftrightarrow$ZH, implemented as optimiser delay.

\paragraph{Results summary}
Official automatic evaluation results for all final systems on the WMT19 test set are summarised in Table~\ref{tab:results-summary}. Throughout the paper, \bleu is calculated using \sacrebleu\footnote{\url{https://github.com/mjpost/sacreBLEU}} \citep{post-2018-call} unless otherwise indicated.  Our final EN-GU models are available for download.\footnote{See \url{data.statmt.org/wmt19_systems/} for our released EN-GU models and running scripts.}$^,$\footnote{Note that following the discovery of a pre-processing error, the EN$\rightarrow$GU and GU$\rightarrow$EN models have been retrained and achieve \bleu scores of 16.3 and 22.3 respectively.}

\begin{table}[h]
    \centering\small
    \begin{tabular}{lrr}
    \toprule
    Lang. direction & \bleu & Ranking \\
    \midrule
      EN$\rightarrow$GU & 16.4 & 1 \\
      GU$\rightarrow$EN & 21.4 & 2 \\
      EN$\rightarrow$ZH & 34.4 & 7 \\
      ZH$\rightarrow$EN & 27.7 & 6 \\
      DE$\rightarrow$EN & 35.0 & 9 \\
      EN$\rightarrow$CS & 27.9 & 3 \\
    \bottomrule
    \end{tabular}
    \caption{\label{tab:results-summary} Final \bleu score results and system rankings amongst constrained systems according to automatic evaluation metrics.}
\end{table}

\section{Gujarati $\leftrightarrow$ English}

One of the main challenges for translation between English$\leftrightarrow$Gujarati is that it is a low-resource language pair; there is little openly available parallel data and much of this data is domain-specific and/or noisy (cf.~Section~\ref{sec:guen-data}).
Our aim was therefore to experiment how additional available data can help us to improve translation quality: large quantities of monolingual text for both English and Gujarati, and resources from Hindi (a language related to Gujarati) in the form of monolingual Hindi data and a parallel Hindi-English corpus. We applied semi-supervised translation, backtranslation and pivoting techniques to create a large synthetic parallel corpus from these resources (Section~\ref{sec:guen-synthetic-data}), which we used to augment the small available parallel training corpus, enabling us to train our final supervised MT models (Section~\ref{sec:guen-sup-mt}).

\subsection{Data and pre-processing}\label{sec:guen-data}

We trained our models using only data listed for the task (cf.~Table~\ref{tab:engu-data-para}). Note that we did not have access to the  corpora provided by the Technology Development for Indian Languages Programme, as they were only available to Indian citizens.

\begin{table}[!ht]
    \small
    \centering
    \begin{tabular}{llrr}
    \toprule
        Lang(s) & Corpus & \#sents & Ave. len.\\
        \midrule    
        \multicolumn{4}{c}{\textit{Parallel data}}\\
        \midrule
        EN-GU &  Software data & 107,637 & 7.0 \\
              & Wikipedia & 18,033 & 21.1 \\
              & Wiki titles v1 & 11,671 & 2.1 \\
              & Govin & 10,650 & 17.0 \\
              & Bilingual dictionary & 9,979 & 1.5 \\
              & Bible & 7,807 & 26.4\\
              & Emille & 5,083 & 19.1 \\
        GU-HI & Emille & 7,993 & 19.1  \\
        EN-HI & Bombay IIT & 1.4M & 13.4  \\
        \midrule    
        \multicolumn{4}{c}{\textit{Monolingual data}}\\
        \midrule
        EN & News & 200M & 23.6 \\
        GU & Common crawl & 3.7M  & 21.9 \\
           & Emille & 0.9M & 16.6 \\
           & Wiki-dump & 0.4M & 17.7 \\
           & News & 0.2M &  15.4 \\
        HI & Bombay IIT & 45.1M & 18.7 \\
           & News & 23.6M & 17.0 \\
    \bottomrule
    \end{tabular}
    \caption{\label{tab:engu-data-para}EN-GU Parallel training data used. Average length is calculated in number of tokens per sentence. For the parallel corpora, this is calculated for the first language indicated (i.e.~EN, GU, then EN)}
\end{table}

We pre-processed all data using standard scripts from the Moses toolkit \citep{koehn_moses:_2007}: normalisation, tokenisation, cleaning (of training data only, with a maximum sentence length of 80 tokens) and true-casing for English data, using a model trained on all available news data. The Gujarati data was additionally pre-tokenised using the IndicNLP tokeniser\footnote{ \url{anoopkunchukuttan.github.io/indic\_nlp\_library/}} before Moses tokenisation was applied. We also applied subword segmentation using BPE \citep{sennrich_neural_2016}, with joint subword vocabularies. We experimented with different numbers of BPE operations during training.

\subsection{Creation of synthetic parallel data}\label{sec:guen-synthetic-data}

Data augmentation techniques such as backtranslation \citep{sennrich_improving_2016, edunov_understanding_2018}, which can be used to produce additional synthetic parallel data from monolingual data, are standard in MT. However they require a sufficiently good intermediate MT model to produce translations that are of reasonable quality to be useful for training \citep{cong_duy_vu_hoang_iterative_2018}. This is extremely hard to achieve for this language pair. Our preliminary attempt at parallel-only training yielded a very low \bleu score of 7.8 on the GU$\rightarrow$EN development set using a Nematus-trained shallow RNN with heavy regularisation,\footnote{Learning rate: $5\times 10^{-4}$, word dropout \citep{gal2016theoretically}: $0.3$, hidden state and embedding dropout: $0.5$, batch tokens: $1000$, BPE vocabulary threshold $50$, label smoothing: $0.2$.} and similar scores were found for a Moses phrase-based translation system.

Our solution was to train models for the creation of synthetic data that exploit both monolingual and parallel data during training.

\subsubsection{Semi-supervised MT with cross-lingual language model pre-training}\label{sec:guen-xlm}

We followed the unsupervised training approach in \citep{lample_cross-lingual_2019} to train two MT systems, one for EN$\leftrightarrow$GU and a second for HI$\rightarrow$GU.\footnote{We used the code available at \url{https://github.com/facebookresearch/XLM}}
This involves training unsupervised NMT models with an additional supervised MT training step. Initialisation of the models is done by pre-training parameters using a masked language modelling objective as in Bert \citep{devlin_bert:_2019}, individually for each language (MLM, which stands for \textit{masked language modelling}) and/or cross-lingually (TLM, which stands for \textit{translation language modelling}). The TLM objective is the MLM objective applied to the concatenation of parallel sentences. See \citep{lample_cross-lingual_2019} for more details. 

\subsubsection{EN and GU backtranslation}\label{sec:engu-back}

We trained a single MT model for both language directions EN$\rightarrow$GU and GU$\rightarrow$EN using this approach. For pre-training we used all available data in Table~\ref{tab:engu-data-para} (both the parallel and monolingual datasets) with MLM and TLM objectives. The same data was then used to train the semi-supervised MT model, which achieved a \bleu score of 22.1 for GU$\rightarrow$EN and 12.6 for EN$\rightarrow$GU on the dev set (See the first row in Table~\ref{tab:results-engu}). This model was used to backtranslate 7.3M of monolingual English news data into Gujarati and 5.1M monolingual Gujarati sentences into English.\footnote{We were unable to translate all available monolingual data due to time constraints and limits to GPU resources.}

\paragraph{System and training details}
We use default architectures for both pre-training and translation: 6 layers with 8 transformer heads, embedding dimensions of 1024. Training parameters are also as per the default: batch size of 32, dropout and attention dropout of 0.1, Adam optimisation \citep{kingma2014adam} with a learning rate of 0.0001.

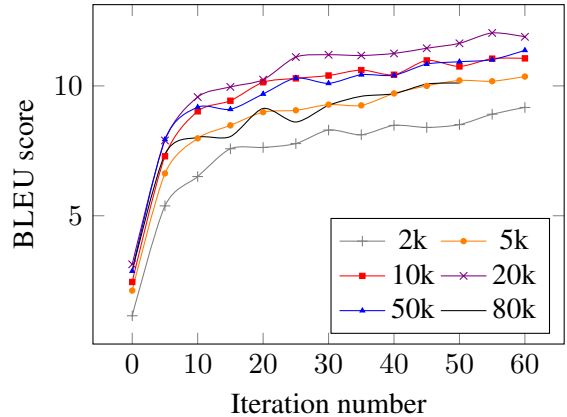
\begin{figure}[!ht]
\begin{tikzpicture}
\pgfplotsset{width=6.2cm,height=4.5cm,compat=1.11}
\begin{axis}[
    scale only axis,
    xlabel={Iteration number},
    ylabel={\bleu score},
    legend pos=south east,
    legend columns=2
]

\addplot[smooth,gray,mark repeat=1,mark=+,mark size=2,each nth point=5]
plot coordinates {
(0,1.15)(1,2.71)(2,3.88)(3,4.6)(4,5.04)(5,5.38)(6,5.43)(7,6.21)(8,6.28)(9,6.29)(10,6.51)(11,6.83)(12,6.82)(13,7.01)(14,7.09)(15,7.58)(16,7.35)(17,7.48)(18,7.38)(19,7.42)(20,7.63)(21,7.76)(22,7.68)(23,7.64)(24,7.76)(25,7.78)(26,7.79)(27,7.55)(28,8.06)(29,7.95)(30,8.3)(31,8.07)(32,7.95)(33,8.08)(34,7.91)(35,8.12)(36,8.25)(37,8.5)(38,8.21)(39,8.32)(40,8.48)(41,8.55)(42,8.33)(43,8.57)(44,8.28)(45,8.4)(46,8.58)(47,8.7)(48,8.62)(49,8.83)(50,8.51)(51,8.89)(52,8.92)(53,8.9)(54,8.87)(55,8.91)(56,8.75)(57,8.94)(58,9.07)(59,8.86)(60,9.17)};

\addplot[smooth,orange,mark repeat=1,mark=*,mark size=1,each nth point=5]
plot coordinates {
(0,2.12)(1,3.87)(2,4.51)(3,5.49)(4,6.04)(5,6.63)(6,7.06)(7,6.99)(8,7.44)(9,7.3)(10,7.98)(11,8.07)(12,7.87)(13,8.22)(14,8.66)(15,8.48)(16,8.55)(17,8.74)(18,8.93)(19,9.13)(20,8.99)(21,8.71)(22,9.08)(23,9.33)(24,9.24)(25,9.06)(26,9.23)(27,9.2)(28,9.36)(29,9.45)(30,9.28)(31,9.27)(32,9.61)(33,9.65)(34,9.69)(35,9.25)(36,9.81)(37,9.68)(38,9.66)(39,9.39)(40,9.71)(41,9.54)(42,9.87)(43,9.71)(44,10.14)(45,10.0)(46,10.0)(47,10.15)(48,9.95)(49,9.88)(50,10.21)(51,10.12)(52,10.33)(53,10.03)(54,9.99)(55,10.18)(56,10.24)(57,10.22)(58,10.19)(59,10.21)(60,10.36)(61,9.96)(62,10.31)};

\addplot[smooth,red,mark=square*, mark size=1,each nth point=5]
plot coordinates {
(0,2.45)(1,4.57)(2,5.75)(3,6.25)(4,6.95)(5,7.29)(6,7.99)(7,7.94)(8,8.32)(9,8.34)(10,9.02)(11,9.41)(12,9.02)(13,9.08)(14,9.09)(15,9.43)(16,9.5)(17,9.65)(18,9.93)(19,9.98)(20,10.14)(21,10.17)(22,10.11)(23,10.5)(24,10.46)(25,10.29)(26,10.03)(27,10.24)(28,10.3)(29,10.32)(30,10.4)(31,10.7)(32,10.23)(33,10.35)(34,10.63)(35,10.61)(36,10.63)(37,10.77)(38,10.8)(39,10.52)(40,10.43)(41,10.38)(42,10.6)(43,10.62)(44,10.61)(45,10.98)(46,10.97)(47,11.0)(48,10.85)(49,11.04)(50,10.75)(51,10.89)(52,10.99)(53,11.16)(54,11.01)(55,11.05)(56,11.17)(57,11.25)(58,10.9)(59,10.92)(60,11.06)(61,11.18)(62,11.24)};

\addplot[smooth,violet,mark=x,each nth point=5]
plot coordinates {
(0,3.13)(1,4.49)(2,6.15)(3,7.19)(4,7.46)(5,7.9)(6,8.1)(7,8.65)(8,8.73)(9,8.99)(10,9.57)(11,9.33)(12,9.56)(13,9.9)(14,10.05)(15,9.96)(16,9.98)(17,10.21)(18,10.49)(19,10.39)(20,10.24)(21,10.56)(22,10.58)(23,10.48)(24,10.75)(25,11.11)(26,10.92)(27,11.01)(28,11.07)(29,11.33)(30,11.2)(31,10.83)(32,11.34)(33,11.04)(34,11.18)(35,11.17)(36,10.94)(37,11.02)(38,11.53)(39,11.45)(40,11.25)(41,11.64)(42,11.65)(43,11.47)(44,11.73)(45,11.45)(46,11.81)(47,11.65)(48,11.61)(49,11.74)(50,11.64)(51,12.13)(52,11.86)(53,11.76)(54,11.57)(55,12.04)(56,11.96)(57,12.0)(58,11.93)(59,11.87)(60,11.89)(61,12.03)(62,11.95)
};

\addplot[smooth,blue,each nth point=5,mark=triangle*, mark size=1]
plot coordinates {
(0,2.87)(1,4.92)(2,5.86)(3,6.36)(4,7.16)(5,7.94)(6,8.03)(7,7.96)(8,8.42)(9,8.59)(10,9.18)(11,8.91)(12,8.98)(13,9.41)(14,9.2)(15,9.1)(16,9.39)(17,9.99)(18,9.55)(19,9.56)(20,9.69)(21,10.05)(22,10.19)(23,10.27)(24,9.9)(25,10.3)(26,10.12)(27,10.17)(28,10.29)(29,10.36)(30,10.1)(31,10.26)(32,10.52)(33,10.48)(34,10.61)(35,10.43)(36,10.37)(37,10.65)(38,10.7)(39,10.68)(40,10.41)(41,10.9)(42,10.6)(43,10.72)(44,10.66)(45,10.84)(46,10.96)(47,10.71)(48,10.94)(49,11.01)(50,10.93)(51,11.14)(52,11.02)(53,11.2)(54,10.94)(55,11.01)(56,11.06)(57,11.13)(58,10.84)(59,11.12)(60,11.37)(61,11.1)(62,11.33)};

\addplot[smooth,black,each nth point=5]
plot coordinates {
(0,2.87)(1,4.79)(2,5.94)(3,6.63)(4,6.58)(5,7.37)(6,7.22)(7,7.76)(8,7.39)(9,7.58)(10,8.02)(11,7.97)(12,8.27)(13,8.17)(14,8.52)(15,8.05)(16,8.87)(17,8.45)(18,8.62)(19,8.67)(20,9.13)(21,8.76)(22,8.95)(23,9.05)(24,9.33)(25,8.61)(26,9.1)(27,9.01)(28,9.14)(29,9.53)(30,9.25)(31,9.55)(32,9.62)(33,9.24)(34,9.64)(35,9.6)(36,9.51)(37,9.63)(38,10.08)(39,9.91)(40,9.7)(41,9.41)(42,9.8)(43,9.95)(44,10.08)(45,10.07)(46,9.94)(47,9.96)(48,9.86)(49,10.04)(50,10.11)(51,10.23)(52,9.81)(53,9.93)};

\legend{2k, 5k, 10k, 20k, 50k, 80k}

\end{axis}
\end{tikzpicture}
\caption{\label{fig:xlm-bpe} The effect of the number of subword operations on \bleu score during training for EN$\rightarrow$GU (calculated on the \textit{newsdev2019} dataset).}
\end{figure}

\paragraph{Degree of subword segmentation}
We tested the impact of varying degrees of subword segmentation on translation quality (See Figure~\ref{fig:xlm-bpe}). Contrary to our expectation that a higher degree of segmentation (i.e.~with a very small number of merge operations) would produce better results, as is often the case with very low resource pairs, the best tested value was 20k joint BPE operations. The reason for this could be the extremely limited shared vocabulary between the two languages\footnote{Except for occasional Arabic numbers and romanised proper names in Gujarati texts.} 
or that training on large quantities of monolingual data turns the low resource task into a higher one.

\subsubsection{HI$\rightarrow$GU translation}\label{sec:higu-trans}

\paragraph{Transliteration of Hindi to Gujarati script} We first transliterated all of the Hindi characters into Gujarati characters to encourage vocabulary sharing. As there are slightly more Hindi unicode characters than Gujarati,  Hindi characters with no corresponding Gujarati characters and all non-Hindi characters were simply copied across.

\begin{figure*}[ht]
  \centering
    \includegraphics[width=0.7\textwidth]{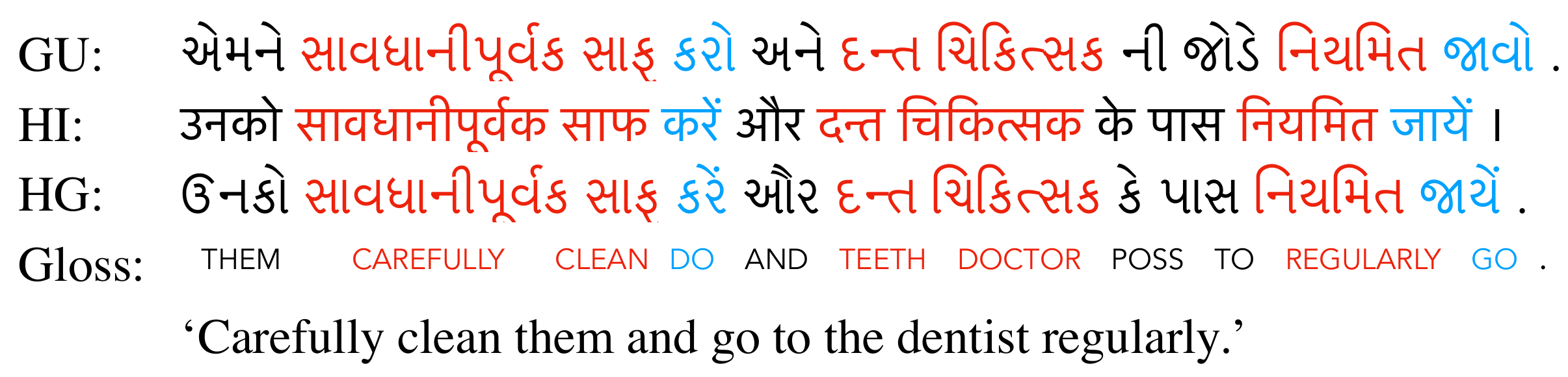}
    \caption{\label{fig:hi2gu-translit} Illustration of Hindi-to-Gujarati transliteration (we refer to the result as HG), with exact matches indicated in red and partial matches in blue.}
\end{figure*}

Once transliterated, there is a high degree of overlap between the transliterated Hindi (HG) and the corresponding Gujarati sentence, which  is demonstrated by the example in Figure~\ref{fig:hi2gu-translit}.

Our parallel Gujarati-Hindi data consisted of approximately 8,000 sentences from the Emille corpus. After transliterating the Hindi, we found that 9\% of Hindi tokens (excluding punctuation and English words) were an exact match to the corresponding Gujarati tokens.
However, we did have access to large quantities of monolingual data in both Gujarati and Hindi (see Table~\ref{tab:engu-data-para}), which we pre-processed in the same way. 

The semi-supervised HI$\leftrightarrow$GU system was trained using the MLM pre-training objective described in Section~\ref{sec:guen-data} and the same model architecture as the EN$\leftrightarrow$GU model in Section~\ref{sec:engu-back}. For the MT step, we trained on 6.5k parallel sentences, reserving the remaining 1.5k as a development set. 
As with the EN$\leftrightarrow$GU model, we investigated the effect of different BPE settings (5k, 10k, 20k and 40k merge operations) on the translation quality. 
Surprisingly, just as with EN$\leftrightarrow$GU, 20k BPE operations performed best (cf.~Table~\ref{tab:results-guhi}), and so we used the model trained in this setting to translate the Hindi side of the IIT Bombay English-Hindi Corpus, which we refer to as HI2GU-EN.

\begin{table}[!ht]
    \centering
    \small
    \begin{tabular}{l|rrrr}
    \toprule
        BPE & 5k & 10k & 20k & 40k \\
        \midrule
        \bleu & 15.4 & 16.0 & 16.3 & 14.6 \\
    \bottomrule
    \end{tabular}
    \caption{\label{tab:results-guhi} The influence of number of BPE merge operations on HI$\rightarrow$GU \bleu score measured using \bleu scores on the development set}
\end{table}

\subsubsection{Finalisation of training data}\label{sec:guen-finaldata}

The final training data for each model was the concatenation of this parallel data, the HI2GU-EN translated data and the  back-translated data for that particular translation direction (See Table~\ref{tab:guen-trainingdata}).

All synthetic data was cleaned by filtering out noisy sentences with consecutively repeated characters or tokens. 
As for the genuine parallel data, we choose only to use the following corpora, which contain an average sentence length of 10 tokens or more: Emille, Govin, Wikipedia and the Bible (a total of approximately 40k sentences). All data was pre-processed using FastBPE\footnote{\url{github.com/glample/fastBPE.git}} with 30k BPE merge operations.

\begin{table}[ht]
    \centering\small
    \begin{tabular}{lrr}
        \toprule
                    & \multicolumn{2}{c}{\#sents} \\
        Training data source & EN$\rightarrow$GU & GU$\rightarrow$EN  \\
        \midrule
        Genuine parallel data & 42k & 42k \\
        HI2GU-EN parallel data & 1.1M & 1.1M \\
        Backtranslated monolingual & 4.5M & 7.1M  \\
        \midrule
        Total  & 5.6M & 8.2M\\
        \bottomrule
    \end{tabular}
    \caption{Summary of EN$\rightarrow$GU and GU$\rightarrow$EN training data, once filtering has been applied to synthetic  data.}
    \label{tab:guen-trainingdata}
\end{table}

\subsection{Supervised MT training}\label{sec:guen-sup-mt}

We trained supervised RNN \citep{micelibarone2017} and transformer models \citep{google_att} using the augmented parallel data augmented described in Section~\ref{sec:guen-finaldata}.
For both model types, we train until convergence and then fine-tuned them on the 40k sentences of genuine parallel data, since synthetic parallel data accounted for more than 99\% of total training data in both translation directions. Results are shown in Table~\ref{tab:results-engu}, our final model results being shown in bold.

\subsubsection{RNN}

Our RNN submission was a BiDeep GRU sequence-to-sequence model \citep{micelibarone2017} with multi-head attention.
The implementation and configuration are the same as in our submission to WMT 2018 \citep{wmt18_edin}, except that we use $1$ attention hop with $4$ attention heads, with a linear projection to dimension $256$ followed by layer normalisation. Other model hyperparameters are encoder and decoder stacking depth: $2$, encoder transition depth: $2$, decoder base level transition depth: $4$, decoder second level transition depth: $2$, embedding dimension: $512$, hidden state dimension: $1024$.
Training is performed with Adam in synchronous SGD mode with initial learning rate: $3 \times 10^{-4}$, label smoothing $0.1$, attention dropout $0.1$ and hidden state dropout $0.1$.
For the final fine-tuning on parallel data we increase the learning rate to $9 \times 10^{-4}$ and hidden state dropout to $0.4$ in order to reduce over-fitting.

\subsubsection{Transformer}\label{sec:guen-transformer}

We trained \textbf{transformer base} models as defined in \citep{google_att}, consisting of 6 encoder layers, 6 decoder layers, 8 heads, with a model/embedding dimension of 512 and feed-forward network dimension of 2048.

We used synchronous SGD, a learning rate of $3 \times 10^{-4}$ and a learning rate warm-up of 16,000. We used a transformer dropout of 0.1.

Our final primary systems are ensembles of four transformers, trained using different random seed initialisations. We also experimented with adjusting the weighting of the models,\footnote{The weights for EN$\rightarrow$GU the  were manually chosen guided  by the individual \bleu scores of the models.} providing gains for EN$\rightarrow$GU but not for GU$\rightarrow$EN, for which equal weighting provided the best results. Our final translations are produced using a beam of 12 for EN$\rightarrow$GU and 60 for GU$\rightarrow$EN.

\subsection{Experiments and results}\label{sec:results-engu}

We report results in Table~\ref{tab:results-engu} on the official development set (1998  sentences) and on the official test sets (998 sentences for EN$\rightarrow$GU and 1016 sentences for GU$\rightarrow$EN). Our results indicate that both the additional synthetic data as well as fine-tuning provide a significant boost in \bleu.

\begin{table}[!ht]
    \centering
    \small
    \begin{tabular}{lrrrr}
    \toprule
           & \multicolumn{2}{c}{EN$\rightarrow$GU} & \multicolumn{2}{c}{GU$\rightarrow$EN} \\
    System & Dev & Test & Dev & Test \\
    \midrule
    Semi-sup. & 12.6 & 11.8 & 22.1 & 15.5  \\
    \midrule 
     RNN & \\
     \quad+ synth. data & 14.2 & 11.4 & 23.4 & 14.7\\
     \quad+ fine-tuning & 15.2 & 11.7 & 24.3 & 15.7\\
    \midrule
    Transformer  &  \\
     \quad+ synth. data & 15.0 & 14.3 & 23.8 & 18.6  \\
     \quad+ fine-tuning & 16.9 & 15.1 & 25.9 & 20.6 \\
     \quad+ Ensemble-4 & 17.9 & 16.5 & 27.2 & \textbf{21.4} \\
     \quad+ Weighted Ensemble & 18.1 & \textbf{16.4} & - & - \\
    \bottomrule
    \end{tabular}
    \caption{\label{tab:results-engu}\bleu scores on the development and test sets for EN$\rightarrow$GU. Our final submissions are marked in bold. Synthetic data is the HI2GU-EN corpus plus backtranslated data for that translation direction and fine-tuning is performed on 40k sentences of genuine parallel data.}
\end{table}

\section{Chinese $\leftrightarrow$ English}

Chinese$\leftrightarrow$English is a high resource language pair with 23.5M sentences of parallel data. The language pair also benefits from a large amount of monolingual data, although compared to English, there is relatively little in-domain (i.e.~news) data for Chinese. Our aim for this year's submission was to test the use of character-based segmentation of Chinese compared to standard subword segmentation, exploiting the properties of the Chinese writing system.

\subsection{Data and pre-processing}
For ZH$\leftrightarrow$EN we pre-processed the parallel data,  which consists of NewsCommentary v13, UN data and CWMT, as follows. The Chinese side of the original parallel data is inconsistently segmented across different corpora so in order to get a consistent segmentation, we desegmented all the Chinese data and resegmented it using the Jieba tokeniser with the default dictionary.\footnote{\url{https://github.com/fxsjy/jieba}} We then removed any sentences that did not contain Chinese characters on the Chinese side or contained only Chinese characters on the English side. We also cleaned up all sentences containing links, sentences longer than 50 words, as well as sentences in which the number of tokens on either side was $>1.3$ times the number of tokens on the other side, following \citet{wmt18_edin}. After pre-processing, the corpus size was 23.6M sentences. We applied BPE with 32,000 merge operations to the English side of the corpora and then removed any tokens appearing fewer than 10 times (which were mostly noise), ending up with a vocabulary size of 32,626. For the Chinese side we attempted two different strategies: A character-level BPE model and a word-level BPE model.

\paragraph*{Character-level Chinese}
A Chinese character-level model is not the same as an English character level model, as it is relatively common for Chinese characters to represent whole words by themselves (in the PKU corpus used for the 2005 Chinese segmentation bakeoff \citep{emerson_second_2005}, a Chinese word contains on average 1.6 characters).
As such, a Chinese character-level model is much more similar to using a BPE model with very few merge operations on English. We hypothesised that using raw Chinese characters in tokenised text makes sense as they form natural subword units.

We segmented all Chinese sentences into characters, but kept non-Chinese characters unsegmented in order to allow for English words and numbers to be kept together as individual units. We then applied BPE with 1,000 merges, which splits the English words in the corpora into mostly trigrams and numbers as bigrams. From the resulting vocabulary we dropped characters occurring fewer than 10 times, resulting in a vocabulary of size 8,535. 

We found that this segmentation strategy was successful for translating into Chinese, however produces significantly worse results when translating from Chinese into English.

\paragraph*{Word-level Chinese}
For word-level Chinese, we took the traditional approach to Chinese pre-processing, where we applied BPE on top of the tokenised dataset. We used 33,000 merge operations and removed tokens occurring fewer than 10 times, resulting in a vocabulary size of 44,529.

\subsection{Iterative backtranslation}
We augmented our parallel data with the same backtranslated ZH$\leftrightarrow$EN as used in \citet{uedin-nmt:2017}, which consists of 8.6M sentences for EN$\rightarrow$ZH from LDC and 9.7M sentences taken from Newscrawl for ZH$\rightarrow$EN. After training the initial systems, we added more backtranslations for both language pairs. For the Chinese side, we used Newscrawl (2.1M sentences) as well as a retranslation of a section of LDC, ending up with 9.5M sentences. For the English side we translated an additional section of Newscrawl, ending up 38M sentences in total. Much to our disappointment, we found that the extra backtranslation is not very effective at increasing the BLEU score, likely because we did not perform any specific domain adaptation for the news domain.

\subsection{Architecture}\label{sec:architecture-zhen}
We used the transformer architecture and three separate configurations.

\paragraph*{Transformer-base}
This is the same architecture as described in Section~\ref{sec:guen-transformer}.

\paragraph*{Transformer-big}\label{para:transformer-big}
6 encoder layers, 6 decoder layers decoder, 16 heads, a model/embedding dimension of 1024, a feedforward network dimension of 4096 and a dropout of 0.1. For character-level Chinese, the number of layers was increased to 8 on the Chinese side. We found transformer-big to be quite fiddly to  train and requires significant hyperparameter exploration. Unfortunately we were unable to find hyperparameters that work effectively for the ZH-EN direction.

\paragraph*{Transfomer-base with larger feed-forward network}
We test \citeauthor{niutrans_zh}'s (\citeyear{niutrans_zh}) recommendation to use the base transformer architecture and increase the feed-forward network (FFNN) size to 4096 instead of using a transformer-big model.

\paragraph*{Ultra-large mini-batches}\label{megabatch}
We follow \citeauthor{ultra_large_minibatch}'s (\citeyear{ultra_large_minibatch}) recommendation to dramatically increase the mini-batch size towards the end of training in order to improve convergence.\footnote{We thank
  Elena Voita for alerting us to this work.} Once our model stopped improving on the development set, we increased the mini-batch size 50-fold by delaying the gradient update \citep{optimizer_delay} to avoid running into memory issues. This increases the average mini-batch size to 13,500 words.

\subsection{Results}
We identified the best single system for each language direction (Tables~\ref{en_zh} and \ref{zh_en}) and ensembled four models trained separately using different random seeds. We also trained right-to-left models, but they got lower scores on the development set and also did not seem to help with ensembling. Our final submission to the competition achieved 28.9 for ZH$\rightarrow$EN and 34.4 for EN$\rightarrow$ZH.

\begin{table}[t]
    \centering\small
    \begin{tabular}{lc}
    \toprule
        System & BLEU \\
         \midrule
         \multicolumn{2}{c}{\textit{Word-level segmentation for ZH}} \\
         \midrule
        Transformer-base & 34.8 \\
        \midrule
        \multicolumn{2}{c}{\textit{Character-level segmentation
        for ZH}} \\
        \midrule
        Transformer-base & 35.1 \\
        \quad + Larger FFNN & 35.6 \\
        Transformer-big & 35.7 \\
        \quad + Ultra-large mini-batches & 36.1 \\
        
        \bottomrule
    \end{tabular}
    \caption{EN$\rightarrow$ZH results on the development set.} 
    \label{en_zh}
\end{table}

\begin{table}[t]
    \centering\small
    \begin{tabular}{lc}
    \toprule
        System & BLEU \\
         \midrule
         \multicolumn{2}{c}{\textit{Word-level segmentation for ZH}} \\
         \midrule
        Transformer-base & 24.1 \\
        \quad + Larger FFNN & 23.7 \\
        \quad \quad + Ultra-large mini-batches & 24.4 \\
        \quad + Ultra-large mini-batches & 24.2 \\
        Transformer-big & 11.3 \\
        \midrule
        \multicolumn{2}{c}{\textit{Character-level segmentation
        for ZH}} \\
        \midrule
        Transformer-base & 20.4 \\
        \bottomrule
    \end{tabular}
    \caption{ZH$\rightarrow$EN results on the development set.} 
    \label{zh_en}
\end{table}

\section{German $\rightarrow$ English}

\begin{table*}
  
  \small\centering
    \begin{tabular}
      {@{\extracolsep{\fill}}llrrr} %
      \toprule
    \bfseries Corpus & \bfseries Type & \bfseries \# of sent. pairs & \bfseries \# of tokens$^1$ (DE) & \bfseries \# of tokens (EN) \\
    \midrule
    Europarl v9     & parallel & 1.82 M & 48.66 M & 51.15 M \\
    Rapid 2019      & parallel & 1.48 M & 30.56 M & 30.95 M \\
    News Commentary & parallel & 0.33 M &  8.51 M &  8.51 M \\
    \color{black}CommonCrawl$^1$ \\
    \color{black}$\quad$ as distributed & \color{black} parallel & \color{black}2.40 M &\color{black} 56.87 M &\color{black} 60.83 M \\
    \color{black}$\quad$ filtered & \color{black} parallel &\color{black} 0.87 M &\color{black} 19.54 M &\color{black} 20.23 M \\
    \color{black}ParaCrawl v3$^2$ \\
    \color{black}$\quad$ as distributed & \color{black} parallel & \color{black} 31.36 M & \color{black} 596.66 M &\color{black} 630.50 M \\
    \color{black}$\quad$ filtered &\color{black} parallel & \color{black} 16.66 M &\color{black} 328.14 M &\color{black} 343.68 M \\
    News Crawl 2007--2018 & English$^3$ & 199.74 M & 4,764.26 M & 4,805.45 M \\
    \midrule
    \\[-1em]
    \multicolumn{5}{l}{\footnotesize $^1$ continuous sequences of  letters,  digits, or repetitions of the same symbol; otherwise, a single symbol.}\\
    \multicolumn{5}{l}{\footnotesize $^2$ used for fine-tuning but not for training the base models, filtered as described in Section~\ref{deen-sec-filtering}.}\\
    \multicolumn{5}{l}{\footnotesize $^3$ German side obtained by back-translation with a model from our participation in WMT18.}\\
    \bottomrule
  \end{tabular}
  \caption{Training data used for German$\rightarrow$English
    translation.}
  \label{de-en-tab-data}
\end{table*}

Following the success of
\citet{edunov_understanding_2018} in WMT18, we decided to focus on
the use of large amounts of monolingual data in the target language.
In addition, we performed fine tuning on data selected
specifically for the test set prior to translation, similar to the method suggested by
\citet{farajian-etal-2017-multi}, but with data selection for the entire test set 
instead of individual sentences.

\subsection{Approach}
Our approach this year is summarised as follows.
\begin{enumerate}
\item Back-translate all available mono-lingual English NewsCrawl data (after filtering out very long sentences). As
  can be seen in Table~\ref{de-en-tab-data}, the amount of monolingual
  data vastly outweighs the amount of parallel data available.
\item Train multiple systems with different blends of genuine
  parallel, out-of-domain data and back-translated in-domain data.
  We did not use any data from CommonCrawl or Paracrawl to train these base models.
    
\item For a given test set, select suitable training
  data from the pool of all available training data (including
  CommonCrawl and Paracrawl) for fine-tuning, based on $n$-gram overlap
  with the source side of the test set, focusing on rare $n$-grams that
  occur fewer than 50 times in the respective sub-corpus\footnote{For
    practical reasons, we sharded the training data based on
    provenance. In addition, each year of the backtranslated news data
    was treated as a separate sub-corpus.}  of training
  data.
  \item Finally, we translate with an 
   ensemble over several
    check-points of the same training run (best \bleu prior to fine-tuning, fine-tuned, best
    mean cross-entropy per word if different from best \bleu, etc.).
\end{enumerate}

\subsection{Data Preparation}
\subsubsection{Tokenisation Scheme}
For tokenisation and sub-word segmentation, we used SentencePiece\footnote{\url{https://github.com/google/sentencepiece}} \cite{taku2018sentencepiece} with the BPE segmentation scheme and a joint vocabulary of 32,000 items.

\subsection{Back-translation}
We back-translated all of the available English NewsCrawl data using
one of the models from our participation in the WMT18 shared task.

\subsection{Data Filtering}\label{deen-sec-filtering}
The CommonCrawl and ParaCrawl datasets consist of parallel data
automatically extracted from web pages from systematic internet
crawls. These datasets contain considerable amounts of noise and poor
quality data. We used dual conditional cross-entropy filtering
\citep{junczys-dowmunt-2018-microsofts} to rank the data in terms of
estimated translation quality, and only retained data that scored
higher than a threshold determined by cursory inspection of the data
by a competent bilingual at various threshold
levels. Table~\ref{de-en-tab-data} shows the amounts of raw and
filtered data.
For training, we limited the training data to sentence pairs of
at most 120 SentencePiece tokens on either side (source or target).

\subsection{Model Training}
\begin{figure*}
 \centering
  \includegraphics[width=0.8\linewidth]{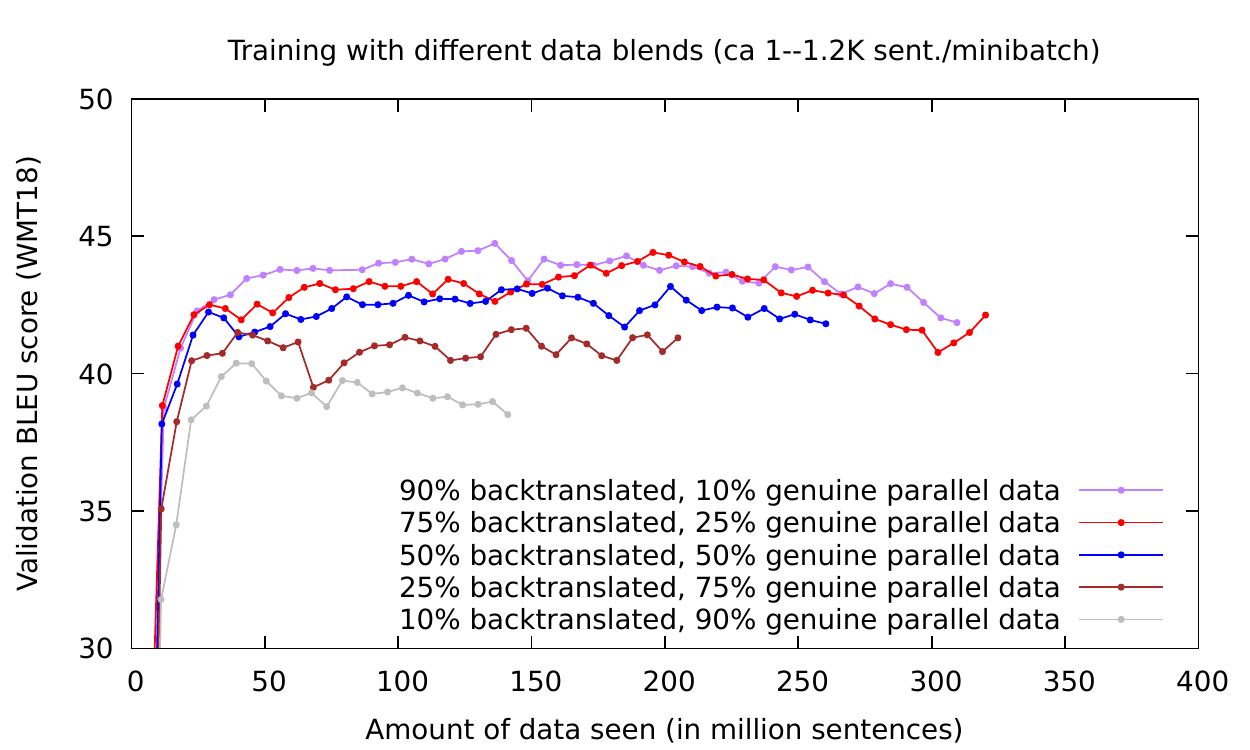}
  \caption{Learning curve for different blends of genuine parallel and synthetic back-translated data. Note that the \bleu scores are inflated with respect to \sacrebleu as they are calculated on BPE-segmented data.}\label{deen-fig-blend}
\end{figure*}

\subsubsection{Initial Training}
To investigate the effect of the blend of genuine parallel and
back-translated news data on translation quality, we trained five
transformer-big models (cf.~Section~\ref{sec:architecture-zhen}) with
different blends of back-translated and genuine parallel data.

We used a dropout value of 0.1 between
transformer layers and no dropout for attention and transformer
filters. We used the Adam optimiser with a learning
rate of 0.0002 and linear warm-up for the first 8K updates, followed by inverted squared decay.

Figure~\ref{deen-fig-blend} shows the learning curves for these
five initial training runs as validated against the WMT18 test set.
Note that the \bleu scores are inflated, as they were computed on the
sub-word units rather than on de-tokenised output. The curves
suggest that adding large amounts of training data does improve
translation quality in direct comparison between the different training
runs. However, compared to last year's top system submissions, these
systems were still lagging behind.

\begin{figure*}
  \centering
  \includegraphics[width=0.8\linewidth,keepaspectratio]{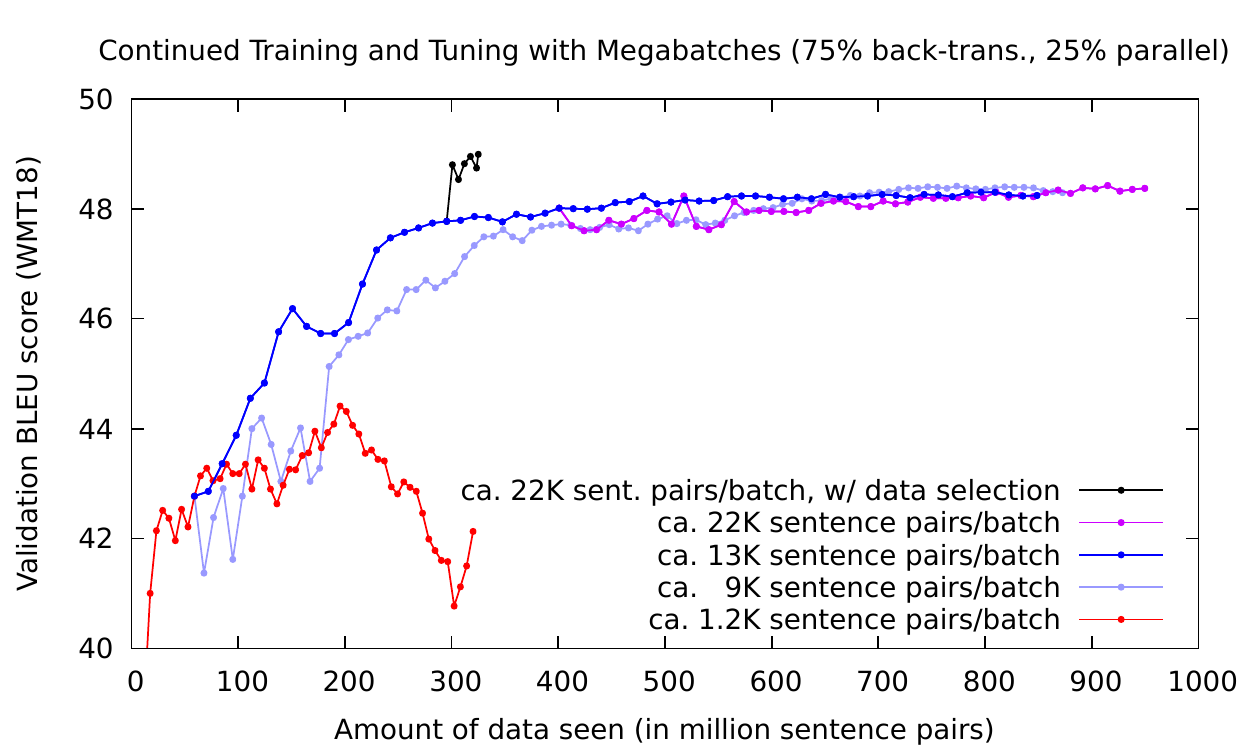}
  \caption{Effect of increased batch size for training and of tuning on data selected for the test set. The red line shows the learning curve for the original training settings (mini-batches of ca. 1,200 sentence pairs). The other lines are the learning curves for models that were initialised with the model parameters of another model at some point in its training process (specifically: at the point where the new learning curve branches off), and then trained with increased batch sizes on the same data (blue and magenta lines), or on data specifically selected to contain rare $n$-grams that also occur in the test / validation set.}\label{deen-fig-megabatch}
\end{figure*}

\subsubsection{Continued training with increased batch size}

Similar to our EN$\leftrightarrow$ZH experiments, we experiment with drastically increasing the mini-batch size by increasing optimiser delay (cf. Section~\ref{megabatch}). Figure~\ref{deen-fig-megabatch} shows the effect of increased mini-batch sizes of ca. 9K, 13K, and 22K sentence pairs,
respectively. The plot shows drastic improvements in the validation scores achieved.

\subsubsection{Fine-tuning on selected data}
As a last step, we selected data specifically for the test set
and continued training on this data for one epoch of this data.
For the WMT18 test set, this gives a significant boost over the starting
point, as the black line in Figure~\ref{deen-fig-megabatch} shows.

\subsection{Results and Analysis}
Due to resource congestion, we were not able to train our models to
convergence in time for submission. The point where the black line
in Figure~\ref{deen-fig-megabatch} branches off shows the state of our
models prior to tuning for a specific test set.

For our submission to the shared task, we ensembled four models:
\begin{itemize}
\item an untuned model trained on a blend of 75\% back-translated data and 25\% genuine parallel data
\item checkpoint models after 500, 2000, and 3000 updates with batches of ca. 13K sentences on data selected specifically for the WMT19 test set. This data included data from CommonCrawl and Paracrawl.
\end{itemize}

With a \bleu score of 36.7 (35.0 cased) --- as opposed to 44.3 (42.8
cased) for the top-performing system --- our results were
disappointing. Apart from a probably suboptimal choice of training
hyperparameters, what else went wrong?

\paragraph{Post-submission analysis}

\begin{table*}[h]
  
  \begin{center}
  \small
    \begin{tabular}{c*{9}{r}}
    \toprule
      &
      & \multicolumn{2}{c}{\bfseries WMT15}
      & \multicolumn{2}{c}{\bfseries WMT16}
      & \multicolumn{2}{c}{\bfseries WMT17}
      & \multicolumn{2}{c}{\bfseries WMT18}\\
      \bfseries System
      & \multicolumn{1}{r}{\bfseries batch\makebox[0pt][l]{$^1$}}
      & \bfseries fwd & \bfseries rev 
      & \bfseries fwd & \bfseries rev 
      & \bfseries fwd & \bfseries rev 
      & \bfseries fwd & \bfseries rev \\\midrule
      10\% back-translated, 90\% parallel
      & 1.2K
      & 20.4 & 34.9 & 27.7 & 44.4 & 25.1 & 37.8 & 28.5 & 46.7\\
      25\% back-translated, 75\% parallel
      & 1.2K
      & 20.0 & 37.7 & 27.5 & 47.5 & 24.9 & 39.8 & 27.5 & 49.4\\
      50\% back-translated, 50\% parallel
      & 1.2K
      & 20.2 & 38.3 & 28.2 & 48.8 & 25.9 & 40.8 & 28.3 & 51.3\\
      75\% back-translated, 25\% parallel
      & 1.2K
      & 20.9 & 39.0 & 29.4 & 49.7 & 26.6 & 41.7 & 29.6 & 52.4\\
      90\% back-translated, 10\% parallel
      & 1.2K
      & 21.2 & 38.6 & 29.0 & 49.6 & 26.8 & 41.5 & 29.7 & 52.8\\
      \midrule
      75\% back-translated, 25\% parallel
      & 1.2K
      & 20.9 & 39.0 & 29.4 & 49.7 & 26.6 & 41.7 & 29.6 & 52.4\\

      75\% back-translated, 25\% parallel
      & 9K
      & 23.2 & 41.2 & 31.8 & 51.8 & 28.7 & 44.2 & 32.6 & 56.3\\  

      75\% back-translated, 25\% parallel
      & 13K
      & 23.2 & 40.9 & 31.8 & 51.3 & 28.6 & 44.1 & 32.4 & 56.2\\
      75\% back-translated, 25\% parallel
      & 22K
      & 23.2 & 41.2 & 31.8 & 51.3 & 28.7 & 44.2 & 32.4 & 56.2\\\hline
      75/25, with tuning for WMT18 & 22K
      & 23.6 & 41.3 & 32.5 & 51.6 & 28.9 & 44.0 & 33.2 & 56.7\\\hline\hline
      \multicolumn{8}{r}{Microsoft Marian 2018 (en$\rightarrow$de)}
      & 52.5 & 41.6\\
      \multicolumn{8}{r}{\citet{edunov_understanding_2018} (en$\rightarrow$de)}
      & 45.8 & 46.1\\\midrule
      \multicolumn{10}{l}{\footnotesize $^1$ batch size in sentence pairs}
    \end{tabular}
  \end{center}
  \caption{Contrastive evaluation (\bleu scores) of performance on genuine
    German~$\rightarrow$~English (fwd) translation vs. English source
    restoration from text originally translated from English into
    German (rev).}
  \label{deen-tab-fwd-vs-rev}
\end{table*}
   In order to understand the effect of
back-translations better, we evaluated our systems on a split of test sets from past years
into ``forward'' (German is the original source language)
and ``reverse'' (the source side of the test set are German
translations of texts originally written in English). The results are
shown in Table~\ref{deen-tab-fwd-vs-rev}. As we can see, most of the
gains from using back-translations are concentrated in the ``reverse''
section of the test sets. The same also holds for
\citeauthor{edunov_understanding_2018}'s (\citeyear{edunov_understanding_2018}) results on the WMT18 test sets for
en$\rightarrow$de.  Notice how it outperforms the top-performing
system (Microsoft Marian) on the reverse translation direction but
lags behind in the forward translation.\footnote{We thank Barry
  Haddow for pointing this out to us and for providing us with the
  split test sets and the split numbers for the Microsoft and Facebook
  systems.}

We see two possible reasons for this phenomenon. The first is that
back-translations produce synthetic data that is closer to the reverse
scenario: translating back from the translation into the source. The
second reason is that the reverse scenario offers a better domain
match: newspapers tend to report relatively more on events and issues
relating to their local audience. A newspaper in Munich will report on
matters relating to Munich; the Los Angeles time will focus on matters
of interest to people living in Southern California.

This became evident when we investigated some strange translation
errors that we observed in our submission to the shared task. For
example, our system often translates ``Münchnerin'' (woman from
Munich) as `miner', `minder', or `mint' and ``Schrebergarten''
(allotment garden) as `shrine' (German: Schrein). When we checked
our back-translated training data for evidence, we noticed that these
are systematic translation errors in our back-translations. While the
word ``Münchnerin'' is frequent in our German data, women from Munich
are rarely mentioned as such in English newspapers. With BPE breaking
up rare words into smaller units, the system learned to translate
``min'' (possibly from ``min$|$t'' (as in the production facility for
coins), which is ``Mün$|$ze'' or ``Mün$|$zprägeanstalt'' in German) into
``Mün''. Once ``Mün'' was chosen in the decoder of the MT system, the
German language model favored the sequence \textit{Mün$|$ch$|$nerin} over \textit{Mün$|$ze} or
the even rarer \textit{Münzprägeanstalt}.

These findings suggest that back-translated data as well needs
curation for domain match and systematic translation errors.

Since this year's test sets consist only of the (more realistic)
``forward'' scenario, we were not able to replicate the gains we
observed for previous test sets when adding more back-translated data.

\section{English $\rightarrow$ Czech}

English-Czech is a high-resource language pair in the WMT News Translation shared task.
For our submission to the EN$\rightarrow$CS track, we investigated the effects of simplifying the data pre-processing and training data filtering, and experimented with larger architectures of the Transformer model.

\subsection{Data and pre-processing}

For English$\rightarrow$Czech experiments we use all parallel corpora available to build a constrained system except CommonCrawl, which is noisy and relatively small compared to the CzEng 1.7 corpus\footnote{\url{https://ufal.mff.cuni.cz/czeng/czeng17}} \cite{czeng16:2016}.
We clean the data following \newcite{popel:2018:WMT} by removing sentence pairs that do not contain at least one Czech diacritic letter.
Duplicated sentences, sentences with $<$3 or $>$200 tokens, and sentences with the ratio of alphabetic to non-alphabetic characters $<$0.5 are also removed.
The final parallel training data contains 44.93M sentences.
For back-translation we use approximately 80M English and Czech monolingual sentences from NewsCrawl \cite{bojar-EtAl:2018:WMT1}, which we cleaned in a similar manner.

\begin{table}[!ht]
\small
\centering
\begin{tabular}{lrrr}
\toprule
    Preprocessing & Dev & 2017 & 2018 \\
\midrule
    Tc + Tok + BPE      & 26.8 & 23.0 & 22.2 \\
    Tc + Tok + ULM      & 26.7 & 22.9 & 22.3 \\
    ULM (raw text)      & 26.7 & 22.9 & \textbf{22.9} \\
    + Resampling        & 26.7 & 22.2 & 21.8 \\
\bottomrule
\end{tabular}
    \caption{Comparison of different pre-processing pipelines for EN$\rightarrow$CS according to \bleu. \textit{Tc} stands for truecasing, \textit{Tok} for tokenisation.}
    \label{encs:seg}
\end{table}

We aimed to explore whether, in a high-resource setting, the common pre- and post-processing pipelines that usually include truecasing, tokenisation and subword segmentation using byte pair encoding (BPE) \cite{sennrich_neural_2016} can be simplified with no loss to performance.
We replace BPE with the segmentation algorithm based on a Unigram Language Model (ULM) from SentencePiece, which is built into Marian.
In both cases we learn 32k subword units jointly on 10M sampled English and Czech sentences.
We gradually remove the elements of the pipeline and find no significant difference between the two segmentation algorithms (Table~\ref{encs:seg}).
We do observe a performance drop when subword resampling is used, but this has been shown to be more effective particularly for Asian languages \cite{taku2018subword}.
For the following English-Czech experiments, we use ULM segmentation on raw text.
    
\subsection{Experiment settings}

We use the transformer-base and transformer-big architectures described in Section~\ref{sec:architecture-zhen}. 
Models are regularised with dropout between transformer layers of 0.2 and in attention of 0.1 and feed-forward layers of 0.1, label smoothing and exponential smoothing: 0.1 and 0.0001 respectively.
We optimise with Adam with a learning rate of 0.0003 and linear warm-up for first 16k updates, followed by inverted squared decay.
For Transformer Big models we decrease the learning rate to 0.0002.
We use mini-batches dynamically fitted into 48GB of GPU memory on 4 GPUs and delay gradient updates to every second iteration, which results in mini-batches of 1-1.2k sentences.
We use early stopping with a patience of 5 based on the word-level cross-entropy on the \textit{newsdev2016} data set.
Each model is validated every 5k updates, and we use the best model checkpoint according to uncased \bleu{} score. 

Decoding is performed with beam search with a beam size of 6 with length normalisation.
Additionally, we reconstruct Czech quotation marks using regular expressions as the only post-processing step \cite{popel:2018:WMT}.

\subsection{Experiments and Results}

\begin{table}[!ht]
\small
\centering
\begin{tabular}{llrrr}
\toprule
    Lang. & System & Dev & 2017 & 2018 \\
\midrule
    \multirow{2}{*}{EN-CS} 
    & Transformer-base                      & 26.7 & 22.9 & 22.9 \\
    & + Data filtering                      & 27.1 & 23.4 & 22.6 \\
\midrule
    \multirow{2}{*}{CS-EN} 
    & Transformer-base                      & 32.6 & 28.8 & 30.3 \\
    & + Back-translation                    & 37.3 & 31.9 & 32.4 \\
\midrule
    \multirow{3}{*}{EN-CS} 
    & Base + Back-transl.                   & 28.4 & 25.1 & 25.1 \\
    & $\rightarrow$ Transformer-big         & 29.6 & 26.3 & 26.2 \\
    & + Ensemble x2                         & 29.6 & 26.5 & 26.3 \\
\bottomrule
\end{tabular}
    \caption{\bleu score results for EN-CS experiments.}
    \label{encs:exp}
\end{table}

Results of our models are shown in Table~\ref{encs:exp}.

We first trained single transformer-base models for each language direction to serve as our baselines.
We then re-score the EN$\rightarrow$CS training data using the CS$\rightarrow$EN model and filter out the 5\% of data with the worst cross-entropy scores, which is a one-directional version of the dual conditional cross-entropy filtering, which we also used for our EN$\rightarrow$DE experiments.
This improves the \bleu scores on the development set and \textit{newstest2017}.
Next, we back-translate English monolingual data and train a CS$\rightarrow$EN model, which in turn is used to generate back-translations for our final systems.
The addition of back-translated data improves the Transformer Base model by 1.7-2.5 \bleu, which is less than the improvement from iterative back-translations reported by \cite{popel:2018:WMT}.
A Transformer Big model trained on the same data is ca. 1.1 \bleu better.

Due to time and resource constraints we train and submit a EN$\rightarrow$CS system (this was the only language direction for English-Czech this year) consisting of just two transformer-big models trained with back-translated data.
Our system achieves 28.3 \bleu on \textit{newstest2019}, 2.1 \bleu less then the top system, which ranks it in third position.

\section{Summary}

This paper reports the experiments run in developing the six systems submitted by the University Edinburgh to the 2019 WMT news translation shared task. Our main contributions have been in different exploitation of additional non-parallel resources, in investigating different pre-processing strategies and in the testing of a variety of NMT training techniques.
We have shown the value of using additional monolingual resources through pre-training and semi-supervised MT for our low-resource language pair EN-GU. For the higher resource language pairs, we also exploit monolingual resources in the form of backtranslation. For GU$\rightarrow$EN in particular we study the effect on translation quality of varying the ratio between between genuine and synthetic parallel training data.
For EN$\rightarrow$ZH, we showed that character-based decoding into Chinese produces better results than the standard subword segmentation approach. In EN$\rightarrow$CS, we also studied the effects of pre-processing, by showing that in such a high resource setting, a simplified pre-processing pipeline can be highly successful. 

Our low resource language pairs, EN$\rightarrow$GU and GU$\rightarrow$EN systems were ranked 1st and 2nd respectively out of the constrained systems according to the automatic evaluation. For the high resource pairs, our EN$\rightarrow$CS system ranked 3rd, EN$\rightarrow$ZH and ZH$\rightarrow$EN ranked 7th and 6th respectively and DE$\rightarrow$EN ranked 9th.

\section*{Acknowledgements}
\lettrine[image=true, lines=2, findent=1ex, nindent=0ex, loversize=.15]{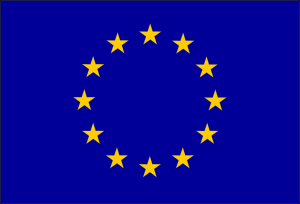}%
{T}his work was supported by funding from the European Union's Horizon 2020 research and innovation programme under grant agreements No 825299 (GoURMET), 825303 (Bergamot), and 825627 (European Language Grid). 

It was also supported by the UK Engineering and Physical Sciences Research Council (EPSRC) fellowship grant EP/S001271/1 (MTStretch).

It was performed using resources provided by the Cambridge Service for Data Driven Discovery (CSD3) operated by the University of Cambridge Research Computing Service (http://www.csd3.cam.ac.uk/), provided by Dell EMC and Intel using Tier-2 funding from the Engineering and Physical Sciences Research Council (capital grant EP/P020259/1), and DiRAC funding from the Science and Technology Facilities Council (www.dirac.ac.uk). 

We express our warmest thanks to Kenneth Heafield, who provided us with access to the computing resources.

\bibliographystyle{acl_natbib}

\bibliography{acl2019}

\begin{thebibliography}{26}
\expandafter\ifx\csname natexlab\endcsname\relax\def\natexlab#1{#1}\fi

\bibitem[{Bogoychev et~al.(2018)Bogoychev, Junczys-Dowmunt, Heafield, and
  Aji}]{optimizer_delay}
Nikolay Bogoychev, Marcin Junczys-Dowmunt, Kenneth Heafield, and Alham~Fikri
  Aji. 2018.
\newblock \href {https://www.aclweb.org/anthology/D18-1332} {Accelerating
  asynchronous stochastic gradient descent for neural machine translation}.
\newblock In \emph{Proceedings of the 2018 Conference on Empirical Methods in
  Natural Language Processing}, EMNLP'18, pages 2991--2996, Brussels, Belgium.

\bibitem[{Bojar et~al.(2016)Bojar, Du{\v{s}}ek, Kocmi, Libovick{\'{y}},
  Nov{\'{a}}k, Popel, Sudarikov, and Vari{\v{s}}}]{czeng16:2016}
Ond{\v{r}}ej Bojar, Ond{\v{r}}ej Du{\v{s}}ek, Tom Kocmi, Jind{\v{r}}ich
  Libovick{\'{y}}, Michal Nov{\'{a}}k, Martin Popel, Roman Sudarikov, and
  Du{\v{s}}an Vari{\v{s}}. 2016.
\newblock {CzEng 1.6: Enlarged Czech-English Parallel Corpus with Processing
  Tools Dockered}.
\newblock In \emph{{Text, Speech, and Dialogue: 19th International Conference,
  {TSD} 2016}}, number 9924 in Lecture Notes in Computer Science, pages
  231--238, Brno, Czech Republic.

\bibitem[{Bojar et~al.(2018)Bojar, Federmann, Fishel, Graham, Haddow, Huck,
  Koehn, and Monz}]{bojar-EtAl:2018:WMT1}
Ondřej Bojar, Christian Federmann, Mark Fishel, Yvette Graham, Barry Haddow,
  Matthias Huck, Philipp Koehn, and Christof Monz. 2018.
\newblock \href {http://www.aclweb.org/anthology/W18-6401} {{Findings of the
  2018 Conference on Machine Translation (WMT18)}}.
\newblock In \emph{Proceedings of the 3rd Conference on Machine Translation,
  Volume 2: Shared Task Papers}, WMT'18, pages 272--307, Belgium, Brussels.

\bibitem[{Devlin et~al.(2019)Devlin, Chang, Lee, and
  Toutanova}]{devlin_bert:_2019}
Jacob Devlin, Ming-Wei Chang, Kenton Lee, and Kristina Toutanova. 2019.
\newblock Bert: {Pre}-training of deep bidirectional transformers for language
  understanding.
\newblock In \emph{Proceedings of the 2019 {Conference} of the {North}
  {American} {Chapter} of the {Association} for {Computational} {Linguistics};
  {Human} {Language} {Technologies}}, {NAACL}-{HLT}'19, Minneapolis, Minnesota.

\bibitem[{Edunov et~al.(2018)Edunov, Ott, Auli, and
  Grangier}]{edunov_understanding_2018}
Sergey Edunov, Myle Ott, Michael Auli, and David Grangier. 2018.
\newblock Understanding {Back}-{Translation} at {Scale}.
\newblock In \emph{Proceedings of the 2018 {Conference} on {Empirical}
  {Methods} in {Natural} {Language} {Processing}}, {EMNLP}'18, pages 489--500,
  Brussels, Belgium.

\bibitem[{Emerson(2005)}]{emerson_second_2005}
Thomas Emerson. 2005.
\newblock \href {https://aclweb.org/anthology/papers/I/I05/I05-3017/} {{The
  Second International Chinese Word Segmentation Bakeoff}}.
\newblock In \emph{Proceedings of the 4th SIGHAN Workshop on Chinese Language
  Processing}, Jeju Island, Korea.

\bibitem[{Farajian et~al.(2017)Farajian, Turchi, Negri, and
  Federico}]{farajian-etal-2017-multi}
M.~Amin Farajian, Marco Turchi, Matteo Negri, and Marcello Federico. 2017.
\newblock \href {https://doi.org/10.18653/v1/W17-4713} {Multi-domain neural
  machine translation through unsupervised adaptation}.
\newblock In \emph{Proceedings of the 2nd Conference on Machine Translation},
  pages 127--137, Copenhagen, Denmark. Association for Computational
  Linguistics.

\bibitem[{Gal and Ghahramani(2016)}]{gal2016theoretically}
Yarin Gal and Zoubin Ghahramani. 2016.
\newblock \href
  {https://papers.nips.cc/paper/6241-a-theoretically-grounded-application-of-dropout-in-recurrent-neural-networks.pdf}
  {A theoretically grounded application of dropout in recurrent neural
  networks}.
\newblock In \emph{{Advances in Neural Information Processing Systems}}, pages
  1019--1027.

\bibitem[{Haddow et~al.(2018)Haddow, Bogoychev, Emelin, Germann, Grundkiewicz,
  Heafield, {Miceli Barone}, and Sennrich}]{wmt18_edin}
Barry Haddow, Nikolay Bogoychev, Denis Emelin, Ulrich Germann, Roman
  Grundkiewicz, Kenneth Heafield, Antonio~Valerio {Miceli Barone}, and Rico
  Sennrich. 2018.
\newblock {{The University of Edinburgh's Submissions to the WMT18 News
  Translation Task}}.
\newblock In \emph{Proceedings of the 3rd Conference on Machine Translation},
  {WMT'18}, pages 399--409, Brussels, Belgium.

\bibitem[{Hoang et~al.(2018)Hoang, Koehn, Haffari, and
  Cohn}]{cong_duy_vu_hoang_iterative_2018}
{Cong Duy Vu} Hoang, Philipp Koehn, Gholamreza Haffari, and Trevor Cohn. 2018.
\newblock Iterative {Back}-{Translation} for {Neural} {Machine} {Translation}.
\newblock In \emph{Proceedings of the 2nd {Workshop} on {Neural} {Machine}
  {Translation} and {Generation}}, {WNMT}'18, pages 18--24, Melbourne,
  Australia.

\bibitem[{Junczys-Dowmunt(2018)}]{junczys-dowmunt-2018-microsofts}
Marcin Junczys-Dowmunt. 2018.
\newblock \href {https://www.aclweb.org/anthology/W18-6415} {{M}icrosoft{'}s
  submission to the {WMT}2018 news translation task: How {I} learned to stop
  worrying and love the data}.
\newblock In \emph{Proceedings of the Third Conference on Machine Translation:
  Shared Task Papers}, pages 425--430, Belgium, Brussels. Association for
  Computational Linguistics.

\bibitem[{Junczys-Dowmunt et~al.(2018)Junczys-Dowmunt, Grundkiewicz, Dwojak,
  Hoang, Heafield, Neckermann, Seide, Germann, Aji, Bogoychev, Martins, and
  Birch}]{junczys-dowmunt_marian:_2018}
Marcin Junczys-Dowmunt, Roman Grundkiewicz, Tomasz Dwojak, Hieu Hoang, Kenneth
  Heafield, Tom Neckermann, Frank Seide, Ulrich Germann, Alham~Fikri Aji,
  Nikolay Bogoychev, André F.~T. Martins, and Alexandra Birch. 2018.
\newblock Marian: {Fast} {Neural} {Machine} {Translation} in {C}++.
\newblock In \emph{Proceedings of the 56th {Annual} {Meeting} of the
  {Association} for {Computational} {Linguistics}}, {ACL}'18, pages 116--121,
  Melbourne, Australia.

\bibitem[{Kingma and Ba(2015)}]{kingma2014adam}
Diederik Kingma and Jimmy Ba. 2015.
\newblock \href {https://arxiv.org/abs/1412.6980} {Adam: A method for
  stochastic optimization}.
\newblock In \emph{Proceedings of the 3rd International Conference on Learning
  Representations}, ICLR'15, San Diego, California, USA.

\bibitem[{Koehn et~al.(2007)Koehn, Hoang, Birch, Callison-Burch, Federico,
  Bertoldi, Cowan, Shen, Moran, Zens, Dyer, Bojar, Constantin, and
  Herbst}]{koehn_moses:_2007}
Philipp Koehn, Hieu Hoang, Alexandra Birch, Chris Callison-Burch, Marcello
  Federico, Nicola Bertoldi, Brooke Cowan, Wade Shen, Christine Moran, Richard
  Zens, Chris Dyer, Ondřej Bojar, Alexandra Constantin, and Evan Herbst. 2007.
\newblock Moses: {Open} {Source} {Toolkit} for {Statistical} {Machine}
  {Translation}.
\newblock In \emph{Proceedings of the 45th {Annual} {Meeting} of the
  {Association} for {Computational} {Linguistics}}, {ACL}'07, pages 177--180,
  Prague, Czech Republic.

\bibitem[{Kudo(2018)}]{taku2018subword}
Taku Kudo. 2018.
\newblock \href {https://www.aclweb.org/anthology/P18-1007} {Subword
  regularization: Improving neural network translation models with multiple
  subword candidates}.
\newblock In \emph{Proceedings of the 56th Annual Meeting of the Association
  for Computational Linguistics (Volume 1: Long Papers)}, pages 66--75,
  Melbourne, Australia. Association for Computational Linguistics.

\bibitem[{Kudo and Richardson(2018)}]{taku2018sentencepiece}
Taku Kudo and John Richardson. 2018.
\newblock \href {https://www.aclweb.org/anthology/D18-2012} {{SentencePiece}: A
  simple and language independent subword tokenizer and detokenizer for neural
  text processing}.
\newblock In \emph{Proceedings of the 2018 Conference on Empirical Methods in
  Natural Language Processing: System Demonstrations}, pages 66--71, Brussels,
  Belgium.

\bibitem[{Lample and Conneau(2019)}]{lample_cross-lingual_2019}
Guillaume Lample and Alexis Conneau. 2019.
\newblock \href {http://arxiv.org/abs/1901.07291} {Cross-lingual {Language}
  {Model} {Pretraining}}.
\newblock In \emph{{arXiv}:1901.07291}.

\bibitem[{{Miceli Barone} et~al.(2017){Miceli Barone}, Helcl, Sennrich, Haddow,
  and Birch}]{micelibarone2017}
Antonio~Valerio {Miceli Barone}, Jind\v{r}ich Helcl, Rico Sennrich, Barry
  Haddow, and Alexandra Birch. 2017.
\newblock \href {https://aclweb.org/anthology/papers/W/W17/W17-4710/} {{Deep
  Architectures for Neural Machine Translation}}.
\newblock In \emph{{Proceedings of the 2nd Conference on Machine Translation,
  Volume 1: Research Papers}}, Copenhagen, Denmark. Association for
  Computational Linguistics.

\bibitem[{Popel(2018)}]{popel:2018:WMT}
Martin Popel. 2018.
\newblock \href {http://www.aclweb.org/anthology/W18-6424} {{CUNI} transformer
  neural mt system for {WMT18}}.
\newblock In \emph{Proceedings of the Third Conference on Machine Translation,
  Volume 2: Shared Task Papers}, pages 486--491, Belgium, Brussels. Association
  for Computational Linguistics.

\bibitem[{Post(2018)}]{post-2018-call}
Matt Post. 2018.
\newblock \href {https://www.aclweb.org/anthology/W18-6319} {A call for clarity
  in reporting {BLEU} scores}.
\newblock In \emph{Proceedings of the 3rd Conference on Machine Translation:
  Research Papers}, pages 186--191, Belgium, Brussels.

\bibitem[{Sennrich et~al.(2017)Sennrich, Birch, Currey, Germann, Haddow,
  Heafield, {Miceli Barone}, and Williams}]{uedin-nmt:2017}
Rico Sennrich, Alexandra Birch, Anna Currey, Ulrich Germann, Barry Haddow,
  Kenneth Heafield, Antonio~Valerio {Miceli Barone}, and Philip Williams. 2017.
\newblock \href {http://www.statmt.org/wmt17/pdf/WMT39.pdf} {{The University of
  Edinburgh's Neural MT Systems for WMT17}}.
\newblock In \emph{{Proceedings of the 2nd Conference on Machine Translation,
  Volume 2: Shared Task Papers}}, pages 389--399, Copenhagen, Denmark.

\bibitem[{Sennrich et~al.(2016{\natexlab{a}})Sennrich, Haddow, and
  Birch}]{sennrich_improving_2016}
Rico Sennrich, Barry Haddow, and Alexandra Birch. 2016{\natexlab{a}}.
\newblock Improving {Neural} {Machine} {Translation} {Models} with
  {Monolingual} {Data}.
\newblock In \emph{Proceedings of the 54th {Annual} {Meeting} of the
  {Association} for {Computational} {Linguistics}}, {ACL}'16, pages 86--96,
  Berlin, Germany.

\bibitem[{Sennrich et~al.(2016{\natexlab{b}})Sennrich, Haddow, and
  Birch}]{sennrich_neural_2016}
Rico Sennrich, Barry Haddow, and Alexandra Birch. 2016{\natexlab{b}}.
\newblock Neural {Machine} {Translation} of {Rare} {Words} with {Subword}
  {Units}.
\newblock In \emph{Proceedings of the 54th {Annual} {Meeting} of the
  {Association} for {Computational} {Linguistics}}, {ACL}'16, pages 1715--1725,
  Berlin, Germany.

\bibitem[{Smith et~al.(2018)Smith, Kindermans, Ying, and
  Le}]{ultra_large_minibatch}
Samuel~L. Smith, Pieter{-}Jan Kindermans, Chris Ying, and Quoc~V. Le. 2018.
\newblock Don't decay the learning rate, increase the batch size.
\newblock In \emph{Proceedings of the 6th International Conference on Learning
  Representations}, ICLR'18, Vancouver, Canada.

\bibitem[{Vaswani et~al.(2017)Vaswani, Shazeer, Parmar, Uszkoreit, Jones,
  Gomez, Kaiser, and Polosukhin}]{google_att}
Ashish Vaswani, Noam Shazeer, Niki Parmar, Jakob Uszkoreit, Llion Jones,
  Aidan~N. Gomez, {\L}ukasz Kaiser, and Illia Polosukhin. 2017.
\newblock \href
  {http://papers.nips.cc/paper/7181-attention-is-all-you-need.pdf} {Attention
  is all you need}.
\newblock In \emph{Advances in Neural Information Processing Systems 30}, pages
  5998--6008. Curran Associates, Inc.

\bibitem[{Wang et~al.(2018)Wang, Li, Liu, Jiang, Zhang, Li, Lin, Xiao, and
  Zhu}]{niutrans_zh}
Qiang Wang, Bei Li, Jiqiang Liu, Bojian Jiang, Zheyang Zhang, Yinqiao Li,
  Ye~Lin, Tong Xiao, and Jingbo Zhu. 2018.
\newblock \href {https://aclanthology.info/papers/W18-6430/w18-6430} {{The
  NiuTrans Machine Translation System for {WMT18}}}.
\newblock In \emph{Proceedings of the 3rd Conference on Machine Translation:
  Shared Task Papers}, WMT'18, pages 528--534, Belgium, Brussels.

\end{thebibliography}

\end{document}